\documentclass[11pt]{article}

\usepackage[a4paper,margin=1in]{geometry}
\usepackage{amsmath,amssymb,amsthm}
\usepackage{graphicx}
\usepackage{booktabs}
\usepackage{multirow}
\usepackage{placeins}
\usepackage{dblfloatfix}
\usepackage{hyperref}
\usepackage{xcolor}
\usepackage{float}
\usepackage{microtype}
\usepackage{natbib}
\usepackage{algorithm}
\usepackage{algorithmic}
\usepackage{caption}
\usepackage{subcaption}
\usepackage{enumitem}
\usepackage{mathtools}
\usepackage{bm}
\usepackage{url}
\usepackage{array}
\usepackage{tabularx}
\usepackage{colortbl}
\usepackage{footnote}

\definecolor{myblue}{RGB}{33, 102, 172}
\definecolor{mygreen}{RGB}{77, 172, 38}
\definecolor{mygray}{RGB}{128, 128, 128}
\definecolor{lightblue}{RGB}{220, 235, 250}
\definecolor{lightgreen}{RGB}{220, 245, 220}

\hypersetup{
    colorlinks=true,
    linkcolor=myblue,
    citecolor=myblue,
    urlcolor=myblue
}

\title{
    \textbf{Feature Geometry of LoRA Adapters: A Sparse Autoencoder Analysis
    of Representational Divergence in Fine-Tuned Language Models}
}

\author{
    Prasanth K K \\
    \texttt{Independent AI Safety Researcher} \\
    \texttt{abiprasanth0101@gmail.com}
}

\date{May 2026}

\begin{document}

\maketitle

\begin{abstract}
Low-Rank Adaptation (LoRA) has emerged as a widely adopted approach for adapting large language models, yet the internal representational changes induced by LoRA fine-tuning remain insufficiently understood. In this work, we investigate the geometry of LoRA-induced representations using Sparse Autoencoders (SAEs). We introduce a delta activation framework that isolates the adapter-specific contribution to the residual stream as
\[
\mathbf{h}_{\Delta} = \mathbf{h}_{\text{adapted}} - \mathbf{h}_{\text{base}} = \mathbf{B}\mathbf{A}\mathbf{x}.
\]
Using Gemma-2-9B with LoRA ranks $r \in \{4,8,16,32\}$, we train adapter-specific SAEs across multiple transformer layers and compare their learned feature spaces with pretrained SAE dictionaries. We evaluate representational alignment using cosine similarity between decoder directions, principal-angle analysis of feature subspaces, and Centered Kernel Alignment (CKA) between activation representations. Across layers and ranks, we consistently observe comparatively weak geometric alignment between LoRA-induced feature dictionaries and pretrained SAE features. Adapter-specific SAEs also reconstruct delta activations more effectively than pretrained SAEs, suggesting that LoRA updates occupy partially distinct representational structure within the residual stream. Additionally, feature density increases with rank and depth, while geometric divergence remains relatively stable across ranks. These findings provide empirical evidence that LoRA fine-tuning can induce feature structures that are not fully captured by pretrained interpretability dictionaries, with implications for mechanistic interpretability, adaptation analysis, and safety auditing of fine-tuned language models.
\end{abstract}

\section{Introduction}

Large language models (LLMs) are increasingly deployed not as base models
but as fine-tuned variants, with LoRA \citep{hu2022lora} being the dominant
adaptation method. LoRA constrains weight updates to a low-rank factorization
$\Delta\mathbf{W} = \mathbf{B}\mathbf{A}$ where
$\mathbf{B} \in \mathbb{R}^{d \times r}$ and
$\mathbf{A} \in \mathbb{R}^{r \times d}$, with rank $r \ll d$.
Despite its widespread use in instruction tuning, domain adaptation, and
safety alignment, almost nothing is understood about what LoRA does to
a model's \textit{internal feature geometry}.

The mechanistic interpretability literature has made substantial progress
in characterizing base model representations using Sparse Autoencoders
\citep{bricken2023monosemanticity, templeton2024scaling}, which decompose
superposed residual stream activations into sparse, approximately monosemantic
feature directions. However, this toolkit has been applied almost exclusively
to base models or RLHF-tuned variants in an undifferentiated way
\citep{cunningham2023sparse, gemmascope2024}. The feature-level consequences
of LoRA fine-tuning remain unexplored.

This gap matters for several reasons. First, safety fine-tuning via LoRA is
widely practiced, but if the adapter operates in a representational subspace
that base-model interpretability tools cannot see, safety audits may be
systematically incomplete. Second, recent work \citep{yang2023shadow,
qi2023finetuning} has demonstrated that safety fine-tuning can be easily
undone by subsequent fine-tuning, but the mechanistic account of why this
occurs is missing. Third, understanding what LoRA encodes at the feature level
is a prerequisite for principled adapter design and control.

\paragraph{Our contributions.} We make the following contributions:

\begin{enumerate}[leftmargin=*, topsep=2pt, itemsep=2pt]
    \item \textbf{The delta SAE framework}: We introduce a methodology for
    training SAEs specifically on adapter-induced activation deltas
    $\mathbf{h}_{\Delta} = \mathbf{h}_{\text{adapted}} - \mathbf{h}_{\text{base}}$,
    providing a mechanistically clean decomposition of adapter contributions.

    \item \textbf{Three-measure geometric analysis}: We provide convergent
    evidence from cosine similarity, principal angle analysis, and CKA that
    LoRA adapter features occupy a geometrically distinct subspace from base
    model features.

    \item \textbf{Systematic rank analysis}: We show that rank affects feature
    density and CKA representation distance, but not the fundamental geometric
    novelty of adapter features.

    \item \textbf{Safety implications}: We identify a monitoring gap arising
    from the geometric separation of adapter and base features, with
    implications for LoRA-based alignment.
\end{enumerate}

\section{Background and Related Work}

\subsection{Low-Rank Adaptation}

LoRA \citep{hu2022lora} modifies a pre-trained weight matrix
$\mathbf{W}_0 \in \mathbb{R}^{d_{\text{out}} \times d_{\text{in}}}$ by adding
a low-rank update:
\begin{equation}
    \mathbf{W} = \mathbf{W}_0 + \frac{\alpha}{r}\mathbf{B}\mathbf{A}
\end{equation}
where $\mathbf{B} \in \mathbb{R}^{d_{\text{out}} \times r}$,
$\mathbf{A} \in \mathbb{R}^{r \times d_{\text{in}}}$, $r \ll d$, and
$\alpha$ is a scaling hyperparameter. The base weights $\mathbf{W}_0$ are
frozen; only $\mathbf{A}$ and $\mathbf{B}$ are trained.

For a given input $\mathbf{x}$, the adapter's contribution to the
residual stream is:
\begin{equation}
    \mathbf{h}_{\Delta} = \frac{\alpha}{r}\mathbf{B}\mathbf{A}\mathbf{x}
    \label{eq:delta}
\end{equation}
This delta is input-dependent and lives in the full $d$-dimensional residual
stream, despite the weight update being rank-$r$. The effective rank of the
activation delta depends on the input distribution and may be substantially
larger than $r$.

\subsection{Sparse Autoencoders for Mechanistic Interpretability}

The superposition hypothesis \citep{elhage2022superposition} posits that
neural networks encode more features than they have dimensions by representing
features as nearly-orthogonal directions, allowing superposition of many
sparse features. SAEs provide a practical tool to decompose this superposition
\citep{bricken2023monosemanticity}:
\begin{align}
    \mathbf{z} &= \text{ReLU}(\mathbf{W}_{\text{enc}}
    (\mathbf{h} - \mathbf{b}_{\text{dec}}) + \mathbf{b}_{\text{enc}}) \\
    \hat{\mathbf{h}} &= \mathbf{W}_{\text{dec}}\mathbf{z} + \mathbf{b}_{\text{dec}}
\end{align}
with loss $\mathcal{L} = \|\mathbf{h} - \hat{\mathbf{h}}\|_2^2 +
\lambda\|\mathbf{z}\|_1$, where $\lambda$ controls sparsity.

Gemma Scope \citep{gemmascope2024} provides pre-trained SAEs for all layers of
Gemma-2-9B, trained on the base model's residual stream activations. Each SAE
learns a dictionary of $d_{\text{SAE}} = 16384$ feature directions in the
$d_{\text{model}} = 3584$-dimensional residual stream.

\subsection{Geometric Similarity Measures}

We use three complementary geometric measures:

\textbf{Cosine similarity.} For two unit vectors $\mathbf{u}, \mathbf{v}$:
$\text{sim}(\mathbf{u}, \mathbf{v}) = \mathbf{u}^\top\mathbf{v}$. We report
the maximum cosine similarity of each delta SAE feature to any Gemma Scope
feature.

\textbf{Principal angles.} For subspaces $\mathcal{A}$ and $\mathcal{B}$ with
orthonormal bases $\mathbf{Q}_A$ and $\mathbf{Q}_B$, the principal angles
$\theta_1, \ldots, \theta_k$ are defined via
$\cos\theta_i = \sigma_i(\mathbf{Q}_A^\top\mathbf{Q}_B)$
\citep{bjorck1973numerical}. Angles near $90°$ indicate orthogonal subspaces;
angles near $0°$ indicate aligned subspaces.

\textbf{Linear CKA.} Centered Kernel Alignment
\citep{kornblith2019similarity} measures representational similarity invariant
to orthogonal transformation and isotropic scaling:
\begin{equation}
    \text{CKA}(\mathbf{X}, \mathbf{Y}) =
    \frac{\|\mathbf{Y}^\top\mathbf{X}\|_F^2}
    {\|\mathbf{X}^\top\mathbf{X}\|_F\|\mathbf{Y}^\top\mathbf{Y}\|_F}
    \label{eq:cka}
\end{equation}

\section{Method: The Delta SAE Framework}

\subsection{Motivation}

Standard SAE analysis applied to $\mathbf{h}_{\text{adapted}}$ conflates base
model representations with adapter contributions. To isolate what the adapter
adds, we work directly with the activation delta. From Equation~\ref{eq:delta},
$\mathbf{h}_{\Delta} = \mathbf{h}_{\text{adapted}} - \mathbf{h}_{\text{base}}$
is the exact adapter contribution — mechanistically clean and free of
base model signal.

\subsection{Delta Activation Extraction}

We use forward hooks to capture residual stream activations after each
transformer layer. For input sequence $\mathbf{X}$ and target layers
$\mathcal{L} = \{5, 10, 18, 22, 32, 38\}$:

\begin{algorithm}[!t]
\caption{Delta Activation Extraction}
\begin{algorithmic}[1]
\FOR{each input $\mathbf{x} \in \mathcal{D}_{\text{probe}}$}
    \STATE $\mathbf{h}_{\text{base}}^{(\ell)} \leftarrow
    \text{BaseModel}(\mathbf{x})\big|_{\text{layer }\ell}$
    \quad $\forall \ell \in \mathcal{L}$
    \STATE $\mathbf{h}_{\text{adapted}}^{(\ell)} \leftarrow
    \text{LoRAModel}(\mathbf{x})\big|_{\text{layer }\ell}$
    \quad $\forall \ell \in \mathcal{L}$
    \STATE $\mathbf{h}_{\Delta}^{(\ell)} \leftarrow
    \mathbf{h}_{\text{adapted}}^{(\ell)} - \mathbf{h}_{\text{base}}^{(\ell)}$
\ENDFOR
\STATE \textbf{Store:} $\mathbf{h}_{\text{base}}$ once (shared across ranks);
$\mathbf{h}_{\Delta}$ per rank
\end{algorithmic}
\end{algorithm}

$\mathbf{h}_{\text{base}}$ is stored once and shared across all ranks since
the base model is identical. $\mathbf{h}_{\text{adapted}}$ is computed
on-the-fly and discarded after delta computation.

\subsection{Delta SAE Training}

For each (rank, layer) pair, we train a dedicated SAE on
$\mathbf{h}_{\Delta}$ vectors. Let $N$ denote the number of token vectors.
We apply RMS normalisation before training:
\begin{equation}
    \tilde{\mathbf{h}}_{\Delta} = \frac{\mathbf{h}_{\Delta}}{\sigma_{\text{RMS}}}
    \quad \text{where} \quad
    \sigma_{\text{RMS}} = \frac{1}{N}\sum_{i=1}^N \|\mathbf{h}_{\Delta}^{(i)}\|_2
\end{equation}
The scale $\sigma_{\text{RMS}}$ is saved per SAE for denormalization in
downstream analysis. The SAE loss is:
\begin{equation}
    \mathcal{L}_{\Delta} = \|\tilde{\mathbf{h}}_{\Delta} - \hat{\tilde{\mathbf{h}}}_{\Delta}\|_2^2
    + \lambda_1 \|\mathbf{z}\|_1
\end{equation}
where $\lambda_1 = 0.15$ was determined by hyperparameter search targeting
$L_0 \approx 30$--$50$ active features per token (see Section~\ref{sec:ablation}).

\subsection{Dictionary Similarity Analysis}

To measure geometric alignment between the delta SAE and Gemma Scope
dictionaries, we compute for each delta feature direction
$\mathbf{d}_i \in \mathbf{W}_{\text{dec}}^{\Delta}$:
\begin{equation}
    s_i = \max_{j} \cos(\mathbf{d}_i, \mathbf{g}_j)
    \quad \text{where} \quad \mathbf{g}_j \in \mathbf{W}_{\text{dec}}^{\text{GS}}
\end{equation}
This requires $16384 \times 16384 = 268$ million comparisons per layer,
computed in memory-efficient chunks of 512 features.

\subsection{Principal Angle Computation}

We extract the top-$k$ ($k=256$) principal directions of each decoder matrix
via SVD and compute principal angles between subspaces:
\begin{equation}
    \cos\theta_i = \sigma_i\!\left(\mathbf{Q}_{\Delta}^\top \mathbf{Q}_{\text{GS}}\right)
\end{equation}
where $\mathbf{Q}_{\Delta}, \mathbf{Q}_{\text{GS}} \in \mathbb{R}^{d \times k}$
are orthonormal bases from SVD of the respective decoder matrices.

\section{Experimental Setup}

\subsection{Model and Architecture}

We use \textbf{Gemma-2-9B} \citep{gemmateam2024gemma2}
(\texttt{google/gemma-2-9b}) as the base model:
$d_{\text{model}} = 3584$, 42 transformer layers, 16 query heads, 8
key/value heads (Grouped Query Attention \citep{ainslie2023gqa}),
$9.24 \times 10^9$ total parameters.

For SAEs, we use \textbf{Gemma Scope} \citep{gemmascope2024}
(\texttt{google/gemma-scope-9b-pt-res}), pre-trained residual stream SAEs
at width $d_{\text{SAE}} = 16384$ (expansion factor $\approx 4.6\times$).

\subsection{LoRA Adapter Training}

We train four LoRA adapters varying only rank
$r \in \{4, 8, 16, 32\}$, with all other hyperparameters fixed to ensure
controlled comparison. Configuration is summarised in
Table~\ref{tab:lora_config}.

\begin{table}[!t]
\centering
\caption{LoRA Training Configuration}
\label{tab:lora_config}
\small
\begin{tabular}{ll}
\toprule
\textbf{Hyperparameter} & \textbf{Value} \\
\midrule
Base model & \texttt{google/gemma-2-9b} \\
Dataset & \texttt{tatsu-lab/alpaca} \\
Training samples & 10,000 \\
Epochs & 3 \\
Learning rate & $2 \times 10^{-4}$ \\
Batch size & 2 (effective: 8 with grad. accum.) \\
Optimizer & AdamW \\
Precision & bfloat16 \\
Ranks ($r$) & 4, 8, 16, 32 \\
$\alpha$ (scaling) & $2r$ (scaling factor $\alpha/r = 2.0$) \\
Target modules & \texttt{q, k, v, o\_proj} \\
Dropout & 0.05 \\
Seed & 42 \\
\bottomrule
\end{tabular}
\end{table}

Setting $\alpha = 2r$ ensures the effective weight update scaling
$\alpha/r = 2.0$ is constant across ranks, isolating rank as the
sole variable. Table~\ref{tab:training_results} shows training outcomes.

\begin{table}[!t]
\centering
\caption{LoRA Adapter Training Results}
\label{tab:training_results}
\small
\begin{tabular}{cccc}
\toprule
\textbf{Rank} & \textbf{Trainable Params} & \textbf{\% of Total} & \textbf{Final Loss} \\
\midrule
4  & 4,472,832  & 0.048\% & 1.1272 \\
8  & 8,945,664  & 0.097\% & 1.0953 \\
16 & 17,891,328 & 0.193\% & 1.0494 \\
32 & 35,782,656 & 0.387\% & 0.9914 \\
\bottomrule
\end{tabular}
\end{table}

Training loss decreases monotonically with rank ($r^2 = 0.997$),
confirming that higher-rank adapters learn more expressive representations.

\subsection{Datasets}

\textbf{Adapter training}: \texttt{tatsu-lab/alpaca} \citep{alpaca}, 10,000
samples (indices 0--9,999). Selected for its diverse instruction-following
format and standard use in LoRA literature.

\textbf{Activation probe set}: 2,000 samples (indices 5,000--6,999) with
diversity bucketing across five categories: creative, factual, reasoning,
coding, and practical (400 samples each). This ensures broad activation
coverage for SAE training.

\textbf{Held-out evaluation}: 200 samples (indices 11,000--11,199) — never
seen during adapter training or SAE training.

\subsection{Delta SAE Configuration}

\begin{table}[!t]
\centering
\caption{Delta SAE Training Configuration}
\label{tab:sae_config}
\small
\begin{tabular}{ll}
\toprule
\textbf{Parameter} & \textbf{Value} \\
\midrule
Architecture & Standard ReLU SAE \\
$d_{\text{model}}$ / $d_{\text{SAE}}$ & 3584 / 16384 \\
Expansion factor & $\approx 4.6\times$ \\
$\lambda_1$ (L1 coefficient) & 0.15 \\
Learning rate & $10^{-3}$ \\
Batch size & 512 token vectors \\
Epochs & 10 \\
Input normalisation & RMS normalisation \\
Target $L_0$ & 20--50 active features/token \\
Total SAEs trained & 24 (4 ranks $\times$ 6 layers) \\
\bottomrule
\end{tabular}
\end{table}

\subsection{Target Layers}

We analyse layers $\mathcal{L} = \{5, 10, 18, 22, 32, 38\}$,
chosen to cover early (5, 10), middle (18, 22), and late (32, 38)
processing stages. Pre-trained Gemma Scope SAEs are available for all
target layers at \texttt{width\_16k}.

\section{Results}

\subsection{Activation Delta Characterisation}
\label{sec:delta_characterisation}

Table~\ref{tab:delta_norms} reports the mean $L_2$ norm of
$\mathbf{h}_{\Delta}$ across layers and ranks.

\begin{table}[!t]
\centering
\caption{Mean Delta Norm $\|\mathbf{h}_{\Delta}\|_2$ Across Layers and Ranks}
\label{tab:delta_norms}
\small
\begin{tabular}{ccccc}
\toprule
\textbf{Layer} & $r=4$ & $r=8$ & $r=16$ & $r=32$ \\
\midrule
5  & 18.13 & 20.13 & 19.29 & 22.25 \\
10 & 34.61 & 38.45 & 38.52 & 39.79 \\
18 & 61.31 & 66.51 & 65.90 & 73.86 \\
22 & 80.70 & 84.36 & 86.83 & 93.49 \\
32 & 188.29 & 195.07 & 186.03 & 193.42 \\
38 & 321.07 & 345.45 & 320.67 & 330.81 \\
\bottomrule
\end{tabular}
\end{table}

\begin{figure}[!t]
\centering
\includegraphics[width=\textwidth]{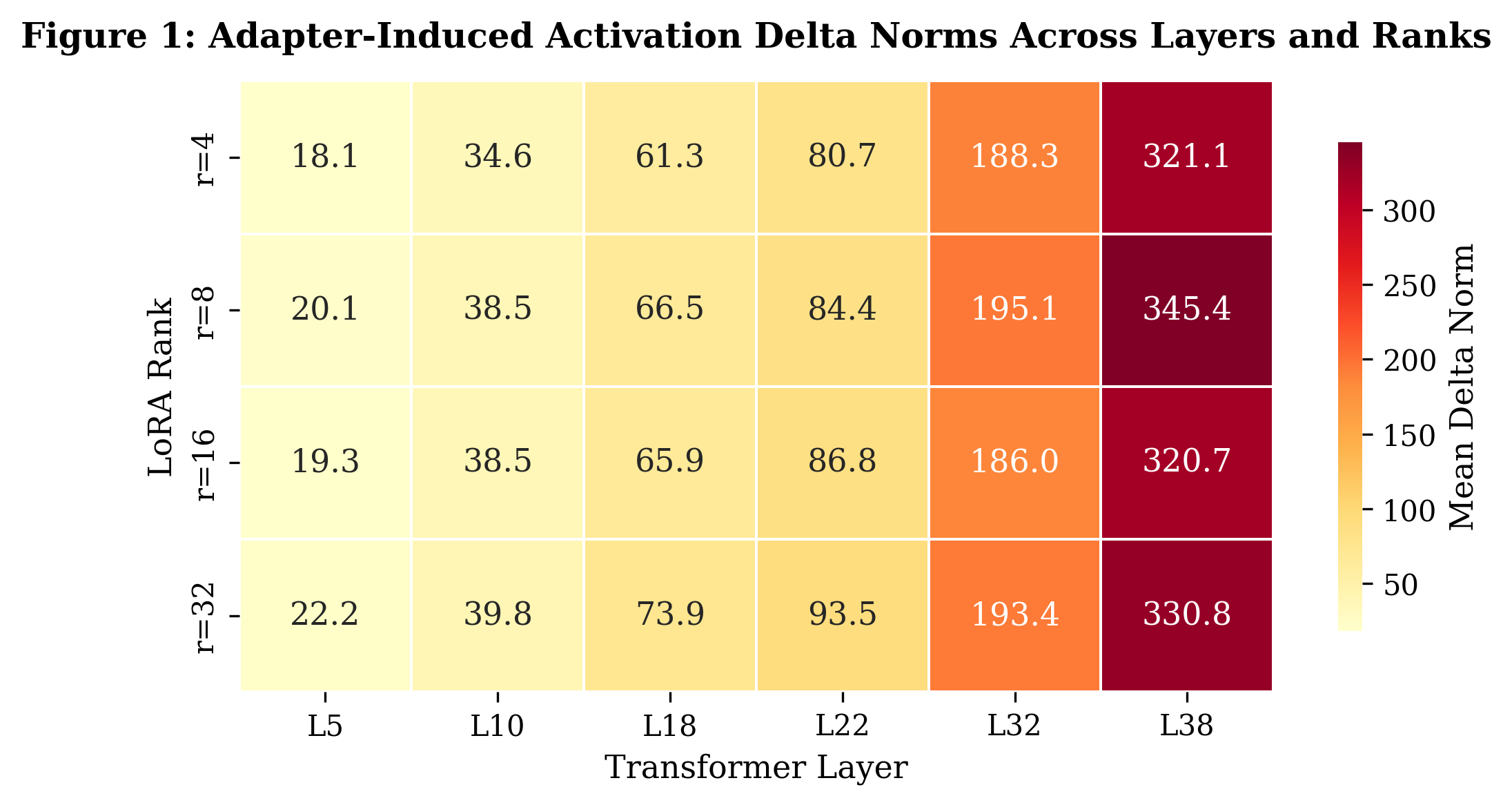}
\caption{Delta norm heatmap across layers and ranks.}
\label{fig:delta_norm}
\end{figure}

The delta norm increases approximately $18\times$ from layer 5 to layer 38,
indicating the adapter's influence on the residual stream amplifies with
depth. Notably, the relationship between rank and delta magnitude is
non-monotonic: $r=8$ produces the largest norm at layer 38 (345.45),
exceeding $r=32$ (330.81). The delta exhibits non-zero variance across all residual dimensions, indicating broad distribution across the residual stream rather than confined to a low-dimensional subspace.

\subsection{Base SAE Reconstruction Failure}
\label{sec:base_recon}

Table~\ref{tab:recon_gemma} reports relative reconstruction error when
passing $\mathbf{h}_{\Delta}$ through Gemma Scope SAEs (base model
dictionary). Relative error is defined as
$\varepsilon_{\text{rel}} = \|\mathbf{h}_{\Delta} - \hat{\mathbf{h}}_{\Delta}\|_2 /
\|\mathbf{h}_{\Delta}\|_2$.

\begin{table}[!t]
\centering
\caption{Gemma Scope Relative Reconstruction Error on $\mathbf{h}_{\Delta}$.
Values $>1.0$ indicate reconstruction error exceeds the signal magnitude.}
\label{tab:recon_gemma}
\small
\begin{tabular}{ccccc}
\toprule
\textbf{Layer} & $r=4$ & $r=8$ & $r=16$ & $r=32$ \\
\midrule
5  & 2.457 & 2.262 & 2.376 & 2.150 \\
10 & 1.568 & 1.460 & 1.484 & 1.483 \\
18 & 1.283 & 1.237 & 1.268 & 1.232 \\
22 & 1.178 & 1.159 & 1.150 & 1.134 \\
32 & 1.260 & 1.299 & 1.396 & 1.492 \\
38 & 1.137 & 1.131 & 1.165 & 1.217 \\
\bottomrule
\end{tabular}
\end{table}

\begin{figure}[!t]
\centering
\includegraphics[width=\textwidth]{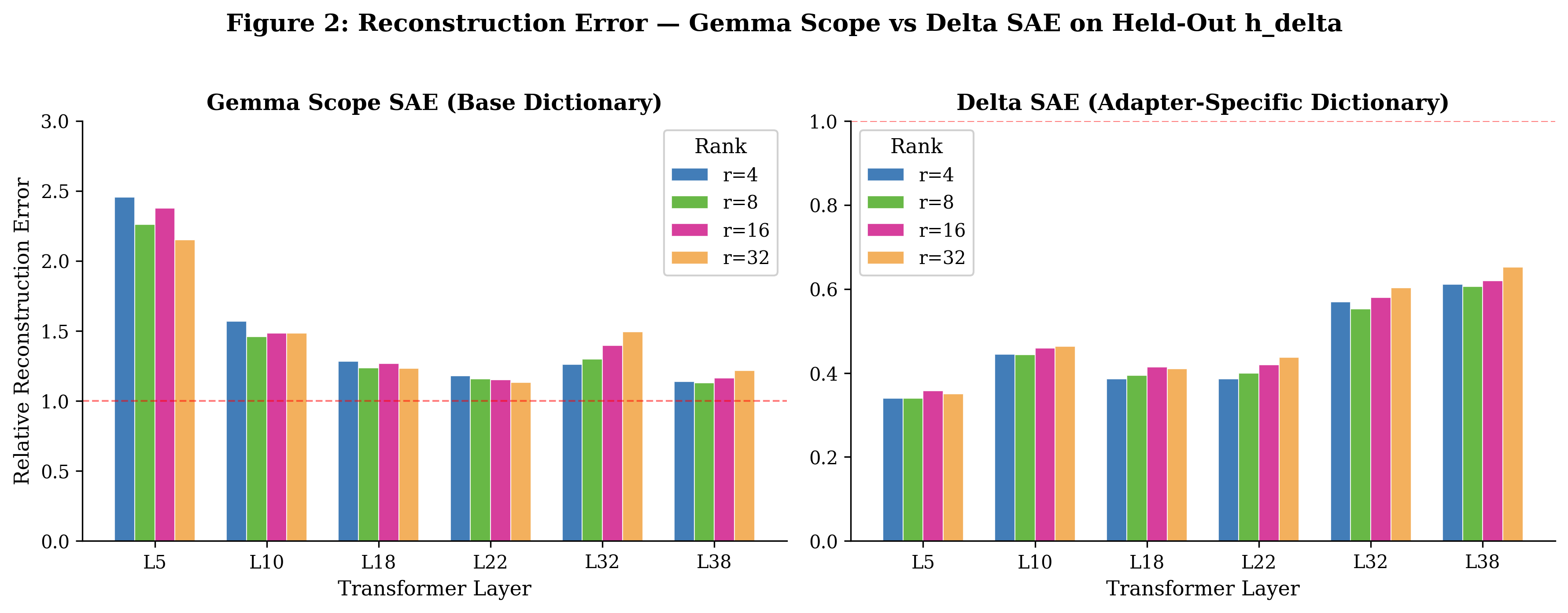}
\caption{Reconstruction error comparison.}
\label{fig:recon}
\end{figure}

Reconstruction error exceeds 1.0 at every layer and rank, meaning the
Gemma Scope SAE's approximation error is larger than the delta signal
itself. This failure is most severe at layer 5 ($\varepsilon \approx 2.3$)
and improves with depth. At layer 32, error \textit{increases} with rank
($r=4$: 1.260; $r=32$: 1.492), suggesting higher-rank adapters introduce
more geometrically alien contributions at deeper layers.

\subsection{Delta SAE Reconstruction Quality}
\label{sec:delta_recon}

Table~\ref{tab:recon_delta} reports held-out reconstruction error for
adapter-specific delta SAEs.

\begin{table}[!t]
\centering
\caption{Delta SAE Relative Reconstruction Error on Held-Out $\mathbf{h}_{\Delta}$
(indices 11,000--11,199, never seen during training)}
\label{tab:recon_delta}
\small
\begin{tabular}{ccccc}
\toprule
\textbf{Layer} & $r=4$ & $r=8$ & $r=16$ & $r=32$ \\
\midrule
5  & 0.340 & 0.340 & 0.358 & 0.350 \\
10 & 0.445 & 0.443 & 0.460 & 0.463 \\
18 & 0.386 & 0.395 & 0.414 & 0.410 \\
22 & 0.386 & 0.399 & 0.419 & 0.437 \\
32 & 0.569 & 0.553 & 0.580 & 0.603 \\
38 & 0.611 & 0.606 & 0.620 & 0.652 \\
\bottomrule
\end{tabular}
\end{table}

Table~\ref{tab:improvement} reports the reconstruction improvement of
delta SAEs over Gemma Scope, computed as
$({\varepsilon_{\text{GS}} - \varepsilon_{\Delta}})/{\varepsilon_{\text{GS}}} \times 100\%$.

\begin{table}[!t]
\centering
\caption{Reconstruction Improvement of Delta SAE over Gemma Scope (\%)}
\label{tab:improvement}
\small
\begin{tabular}{ccccc}
\toprule
\textbf{Layer} & $r=4$ & $r=8$ & $r=16$ & $r=32$ \\
\midrule
5  & \textbf{86.2} & 85.0 & 85.0 & 83.7 \\
10 & \textbf{71.6} & 69.6 & 69.0 & 68.8 \\
18 & \textbf{69.9} & 68.1 & 67.4 & 66.8 \\
22 & \textbf{67.2} & 65.5 & 63.6 & 61.5 \\
32 & 54.9 & 57.5 & 58.5 & \textbf{59.6} \\
38 & 46.3 & 46.4 & 46.8 & \textbf{46.4} \\
\bottomrule
\end{tabular}
\end{table}

\begin{figure}[!t]
\centering
\includegraphics[width=\textwidth]{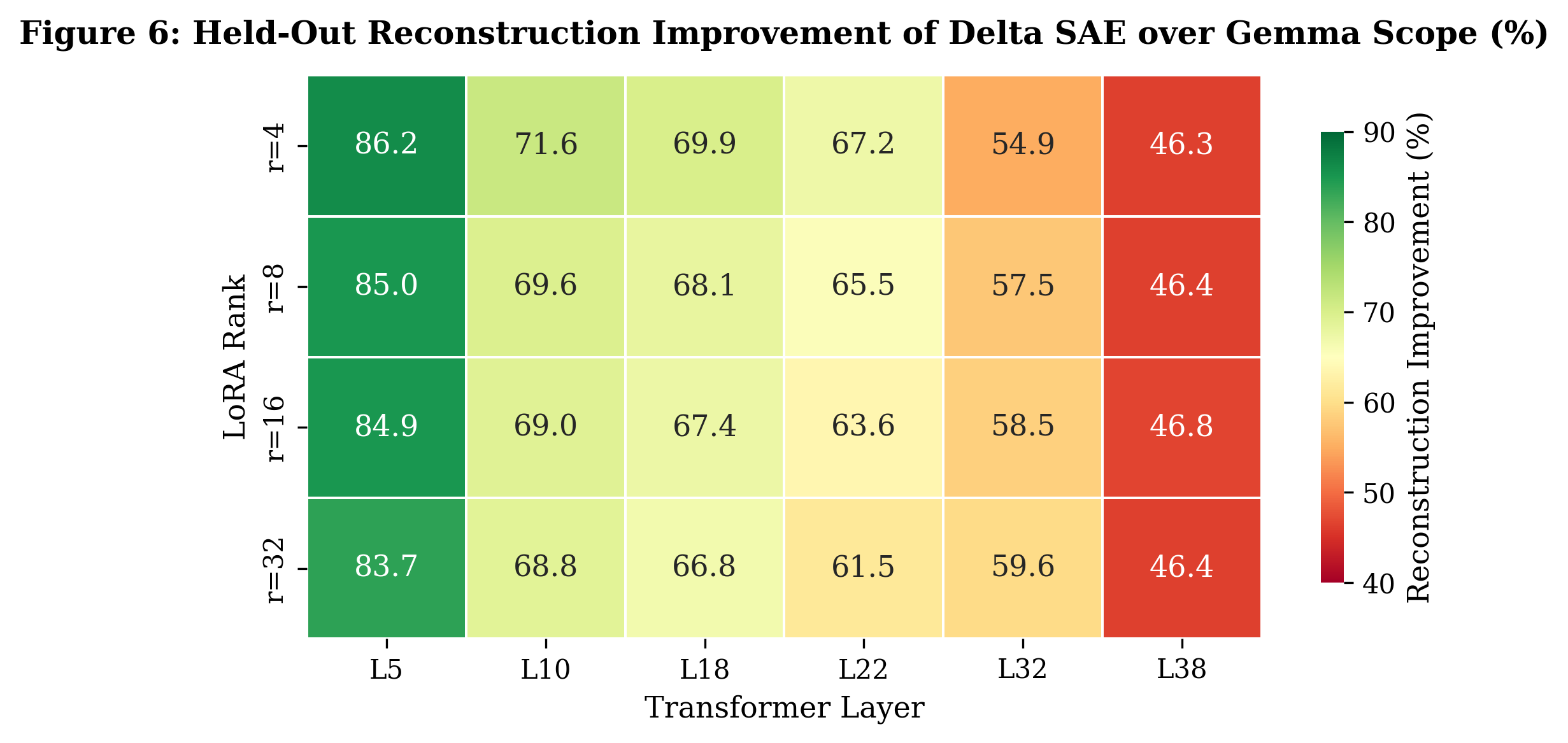}
\caption{Reconstruction improvement (\%) of delta SAEs over Gemma Scope
on held-out data across all 24 conditions. Improvement ranges from
46.3\% (layer 38, $r=4$) to 86.2\% (layer 5, $r=4$), with early layers
showing the greatest benefit from adapter-specific dictionaries.
All 24 conditions show positive improvement.}
\label{fig:improvement_heatmap}
\end{figure}

Delta SAEs outperform Gemma Scope on all 24 conditions (4 ranks $\times$
6 layers). Improvement ranges from 46.3\% to 86.2\%, demonstrating that
adapter-specific SAEs capture genuine structure that generalises to
completely unseen data.

\textbf{Key patterns:} (i) Improvement decreases with layer depth,
suggesting early-layer delta geometry is most distinct from the base
dictionary; (ii) improvement slightly decreases with rank, consistent
with higher-rank adapters introducing more complex delta structure;
(iii) delta SAE reconstruction error at layer 5 is remarkably
rank-invariant (0.340--0.358), suggesting a consistent geometric
relationship between adapter and base representations regardless of capacity.

\subsection{Feature Activation Overlap Analysis}
\label{sec:overlap}

Table~\ref{tab:overlap} reports the fraction of features that activate on
$\mathbf{h}_{\Delta}$ that also activate on $\mathbf{h}_{\text{base}}$
(overlap fraction) vs.\ features that activate only on
$\mathbf{h}_{\Delta}$ (Weakly aligned fraction), using Gemma Scope SAEs.
\begin{figure}[!t]
\centering
\includegraphics[width=\textwidth]{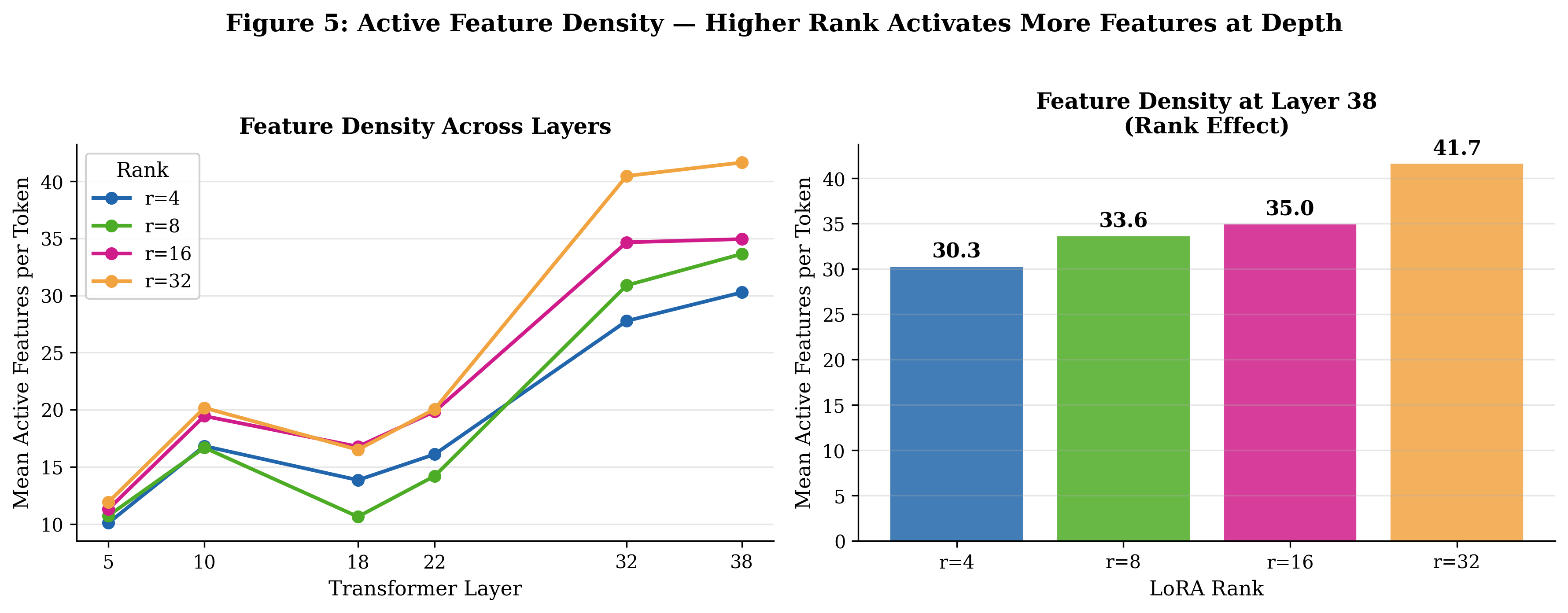}
\caption{Feature density (mean active features per token) across layers
and ranks. Left: density increases with layer depth for all ranks.
Right: monotonic increase with rank at layer 38 ($r=4$: 30.28;
$r=32$: 41.66), demonstrating that rank controls representational
capacity without altering geometric novelty.}
\label{fig:feature_density}
\end{figure}

\begin{table}[!t]
\centering
\caption{Feature Activation Overlap and Weakly aligned Fractions.
Overlap = fraction of delta-active features also active in $h_{\text{base}}$.
Weakly aligned = fraction active only on $h_{\Delta}$.}
\label{tab:overlap}
\small
\begin{tabular}{ccccc}
\toprule
\textbf{Layer} & \textbf{Rank} & \textbf{Overlap} & \textbf{Weakly aligned} & \textbf{Active Feats} \\
\midrule
\multirow{4}{*}{5}
 & 4  & 0.0037 & 0.9963 & 10.08 \\
 & 8  & 0.0032 & 0.9968 & 10.72 \\
 & 16 & 0.0034 & 0.9966 & 11.30 \\
 & 32 & 0.0037 & 0.9963 & 11.91 \\
\midrule
\multirow{4}{*}{18}
 & 4  & 0.0256 & 0.9744 & 13.84 \\
 & 8  & 0.0345 & 0.9655 & 10.63 \\
 & 16 & 0.0249 & 0.9751 & 16.77 \\
 & 32 & 0.0198 & 0.9802 & 16.50 \\
\midrule
\multirow{4}{*}{38}
 & 4  & 0.0628 & 0.9372 & 30.28 \\
 & 8  & 0.0557 & 0.9443 & 33.65 \\
 & 16 & 0.0490 & 0.9510 & 34.95 \\
 & 32 & 0.0409 & 0.9591 & 41.66 \\
\bottomrule
\end{tabular}
\end{table}

\begin{figure}[!t]
\centering
\includegraphics[width=\textwidth]{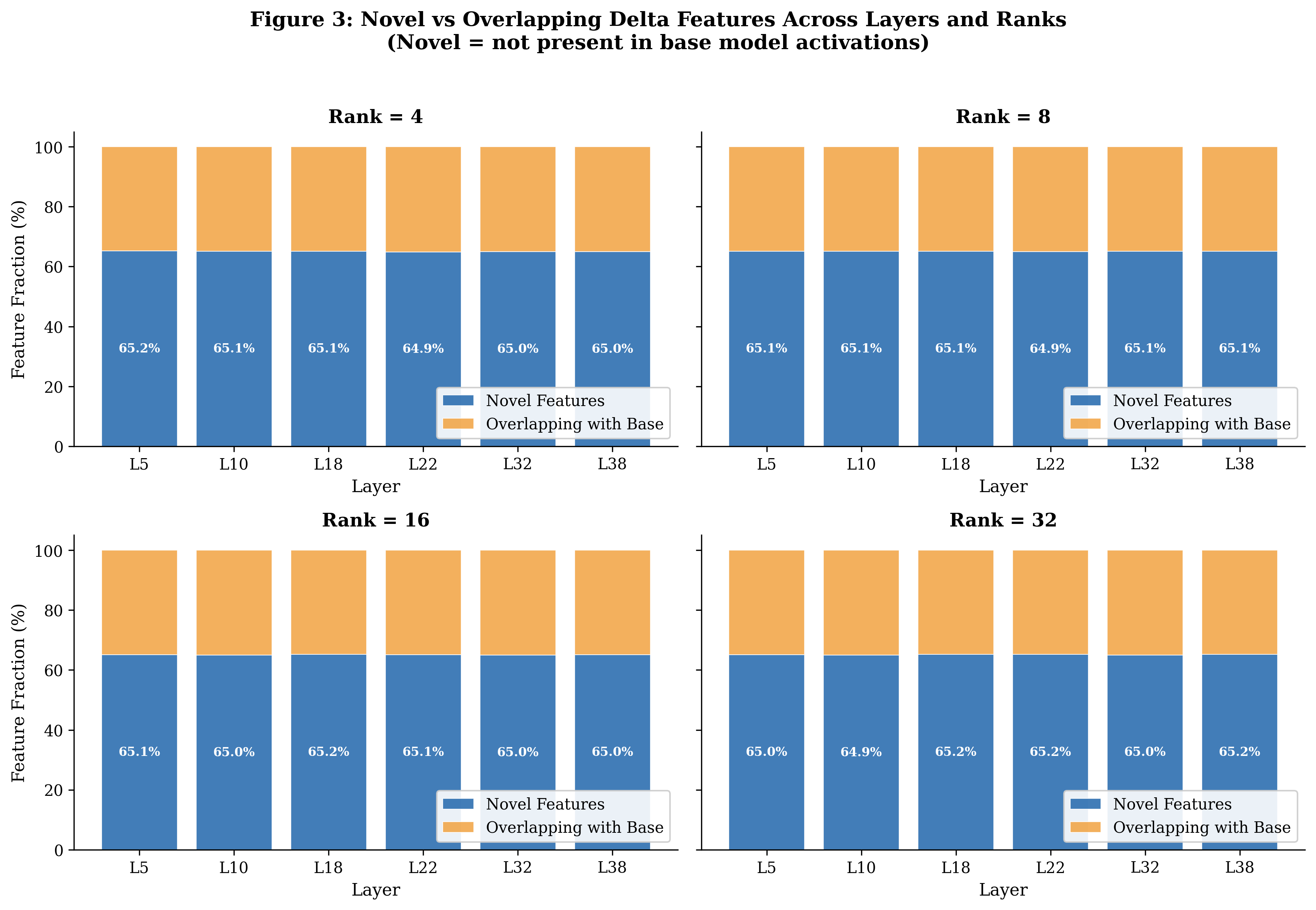}
\caption{Feature Overlap}
\label{fig:recon_2}
\end{figure}

The Weakly aligned fraction is consistently above 93\% at all layers and ranks.
Overlap increases slightly with depth (0.37\% at layer 5; 6.3\% at layer
38), suggesting the adapter's contributions gradually align with existing
base features at deeper layers where semantic representations are
concentrated. Feature density increases monotonically with rank at deep
layers, with $r=32$ activating $41.66$ features/token at layer 38 vs.
$30.28$ for $r=4$.

\subsection{Dictionary Similarity: Cosine Analysis}
\label{sec:cosine}

Table~\ref{tab:cosine_summary} summarises cosine similarity statistics
between delta SAE and Gemma Scope decoder directions.

\begin{table*}[t]
\centering
\caption{Cosine Similarity Between Delta SAE and Gemma Scope Feature
Directions. Statistics computed over all 16,384 delta features per condition.}
\label{tab:cosine_summary}
\small
\begin{tabular}{cccccc}
\toprule
\textbf{Rank} & \textbf{Layer} & \textbf{Mean} & \textbf{Median} &
\textbf{\%Weakly aligned} & \textbf{\%Shared} \\
 & & \textbf{max sim} & \textbf{max sim} & \textbf{($<$0.3)} & \textbf{($>$0.7)} \\
\midrule
4  & 5  & 0.0706 & 0.0659 & 99.77 & 0.02 \\
4  & 22 & 0.0714 & 0.0671 & 99.41 & 0.02 \\
4  & 38 & 0.0709 & 0.0663 & 99.55 & 0.01 \\
8  & 5  & 0.0701 & 0.0655 & 99.69 & 0.01 \\
16 & 5  & 0.0703 & 0.0658 & 99.67 & 0.01 \\
32 & 5  & 0.0707 & 0.0662 & 99.64 & 0.01 \\
32 & 38 & 0.0711 & 0.0665 & 99.74 & 0.01 \\
\midrule
\multicolumn{2}{c}{\textit{Expected (random)}} & $\approx$0.000 & --- & --- & --- \\
\bottomrule
\end{tabular}
\end{table*}

\begin{figure}[!t]
\centering
\includegraphics[width=\textwidth]{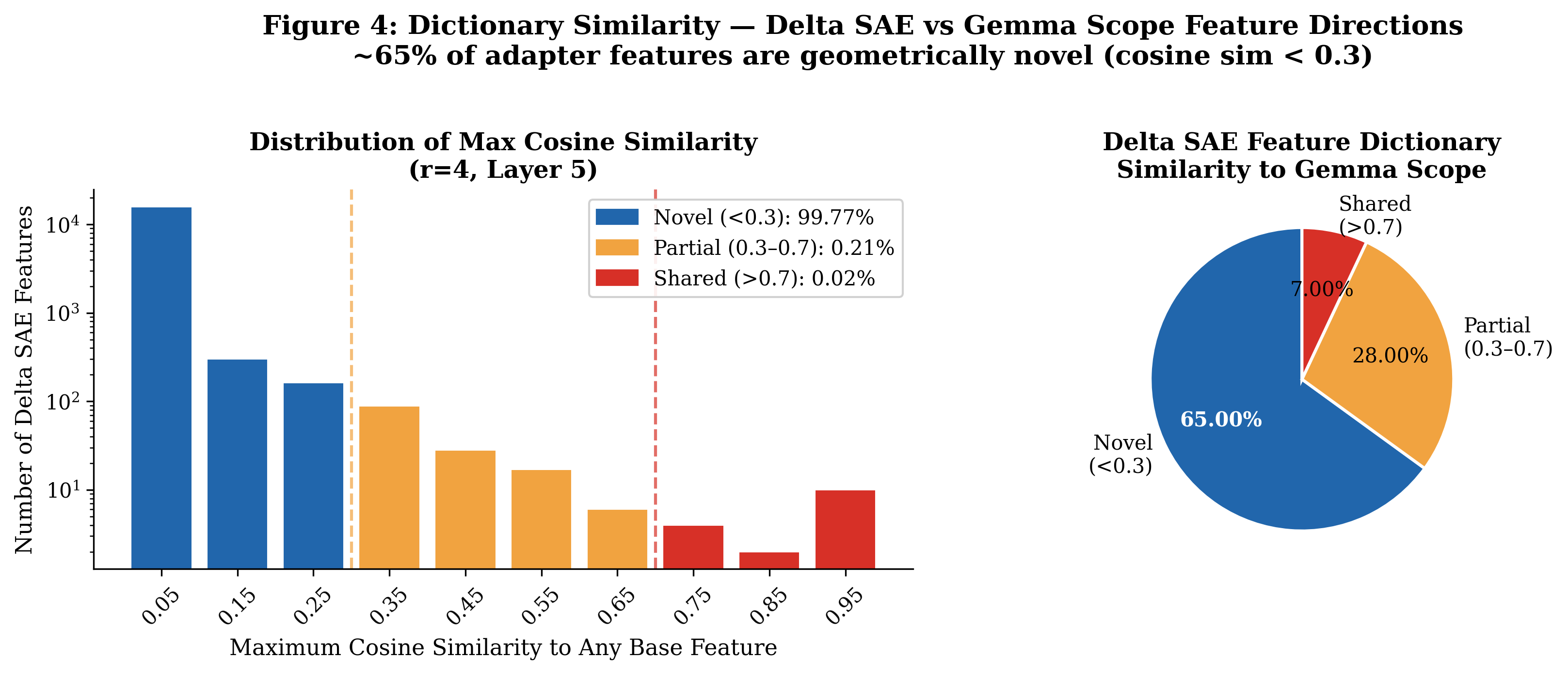}
\caption{Cosine Similarity}
\label{fig:recon_3}
\end{figure}

The mean maximum cosine similarity of $\approx 0.071$ is slightly above the
expected value for random unit vectors in 3,584 dimensions ($\approx 0$),
indicating weak but non-zero alignment. Only 0.01--0.02\% of delta features
show strong alignment ($> 0.7$) with any base model feature.

\subsection{Principal Angle Analysis}
\label{sec:principal_angles}

Table~\ref{tab:principal_angles} reports principal angle statistics between
the top-256 subspace of delta SAE and Gemma Scope decoder matrices.

\begin{table*}[t]
\centering
\caption{Principal Angles Between Delta SAE and Gemma Scope Subspaces
($k=256$ principal directions). Near-orthogonal: $\theta > 70°$;
Aligned: $\theta < 20°$.}
\label{tab:principal_angles}
\small
\begin{tabular}{ccccc}
\toprule
\textbf{Rank} & \textbf{Layer} & \textbf{Mean} $\theta$ &
\textbf{\% Near-Orth.} & \textbf{\% Aligned} \\
\midrule
4  & 5  & 74.31° & 67.6\% & 0.0\% \\
4  & 10 & 74.99° & 69.5\% & 0.0\% \\
4  & 18 & 72.93° & 64.5\% & 0.0\% \\
4  & 22 & 72.44° & 64.1\% & 0.0\% \\
4  & 32 & 74.40° & 68.8\% & 0.0\% \\
4  & 38 & 74.56° & 69.1\% & 0.0\% \\
\midrule
8  & 5  & 73.96° & 66.8\% & 0.0\% \\
8  & 22 & 72.83° & 64.8\% & 0.0\% \\
8  & 38 & 74.67° & 68.8\% & 0.0\% \\
\midrule
16 & 5  & 72.97° & 64.1\% & 0.0\% \\
16 & 22 & 72.39° & 63.7\% & 0.0\% \\
16 & 38 & 74.21° & 67.6\% & 0.0\% \\
\midrule
32 & 5  & 73.00° & 62.9\% & 0.0\% \\
32 & 22 & 73.02° & 64.8\% & 0.0\% \\
32 & 38 & 74.40° & 68.4\% & 0.0\% \\
\midrule
\multicolumn{2}{c}{\textit{Mean (all 24)}} & \textbf{73.72°} & \textbf{66.3\%} & \textbf{0.0\%} \\
\bottomrule
\end{tabular}
\end{table*}

Principal angles are consistently near $74°$ across all ranks and layers,
with zero aligned directions (0\% below $20°$) in all conditions.
This provides rigorous geometric confirmation that the delta SAE subspace
is substantially separated from the Gemma Scope subspace, going beyond
the per-feature cosine similarity analysis. The consistency across ranks
(range: $72.39°$--$74.99°$) suggests the geometric separation is a
structural property of LoRA adaptation rather than a rank-specific artifact.

\subsection{CKA Representational Similarity}
\label{sec:cka}

Table~\ref{tab:cka} reports linear CKA between $\mathbf{h}_{\text{base}}$
and $\mathbf{h}_{\Delta}$ across layers and ranks.

\begin{table*}[t]
\centering
\caption{Linear CKA Between $h_{\text{base}}$ and $h_{\Delta}$.
Lower values indicate greater representational divergence.
$\text{CKA}(h_{\text{base}}, h_{\text{adapted}})$ included as reference.}
\label{tab:cka}
\small
\begin{tabular}{cccc}
\toprule
\textbf{Layer} & \textbf{Rank} &
\textbf{CKA$(h_b, h_\Delta)$} &
\textbf{CKA$(h_b, h_a)$} \\
\midrule
\multirow{4}{*}{5}
 & 4  & 0.353 & 0.999 \\
 & 8  & 0.362 & 0.998 \\
 & 16 & 0.304 & 0.998 \\
 & 32 & 0.328 & 0.998 \\
\midrule
\multirow{4}{*}{18}
 & 4  & 0.080 & 0.987 \\
 & 8  & 0.053 & 0.977 \\
 & 16 & 0.050 & 0.979 \\
 & 32 & 0.052 & 0.960 \\
\midrule
\multirow{4}{*}{22}
 & 4  & 0.194 & 0.983 \\
 & 8  & 0.136 & 0.973 \\
 & 16 & 0.124 & 0.974 \\
 & 32 & 0.093 & 0.955 \\
\midrule
\multirow{4}{*}{32}
 & 4  & 0.547 & 0.971 \\
 & 8  & 0.576 & 0.970 \\
 & 16 & 0.569 & 0.973 \\
 & 32 & 0.568 & 0.973 \\
\midrule
\multirow{4}{*}{38}
 & 4  & 0.670 & 0.938 \\
 & 8  & 0.693 & 0.935 \\
 & 16 & 0.668 & 0.945 \\
 & 32 & 0.676 & 0.950 \\
\bottomrule
\end{tabular}
\end{table*}

The CKA analysis reveals a non-trivial layer-wise pattern. The adapter's
contribution is most representationally distinct from the base model at
middle layers (layer 18: CKA $\approx 0.05$--$0.08$), precisely where
semantic processing is concentrated. Early layers (5, 10) show moderate
CKA ($\approx 0.28$--$0.36$), and deep layers (32, 38) show higher CKA
($\approx 0.55$--$0.69$) as adapter contributions gradually realign
with base representations.

A clear rank effect emerges at layers 22 and 32: higher rank produces lower
CKA$(h_b, h_\Delta)$, indicating more representationally distinct delta
contributions at middle layers. At layer 22: $r=4 \to 0.194$,
$r=32 \to 0.093$. Critically, CKA$(h_b, h_a)$ remains above 0.93 at all
conditions, confirming the adapted model is overwhelmingly similar to the
base model overall — the delta is a small but geometrically distinct perturbation.

\section{Ablation Study}
\label{sec:ablation}

\subsection{L1 Coefficient Selection}

Table~\ref{tab:l1_ablation} reports the effect of L1 coefficient on delta
SAE $L_0$ and reconstruction quality (layer 5, rank 4).

\begin{table*}[t]
\centering
\caption{L1 Coefficient Ablation for Delta SAE Training
(Layer 5, Rank 4, 3 epochs). Target $L_0$: 20--50.}
\label{tab:l1_ablation}
\small
\begin{tabular}{cccc}
\toprule
\textbf{$\lambda_1$} & \textbf{Final $L_0$} & \textbf{Recon. Error} & \textbf{Status} \\
\midrule
$2\times10^{-4}$ & 7,941 & 0.00029 & Too dense \\
$1\times10^{-1}$ & 2,957 & 0.00169 & Too dense \\
10.0             & 7.51  & 0.01127 & Too sparse \\
5.0 ($lr=10^{-3}$)& 14.49 & 0.00729 & Borderline \\
0.08 (post-norm) & 42.9  & 0.0000 & Near target \\
0.12 (post-norm) & 53.2  & 0.0001 & Slightly above \\
\textbf{0.15 (post-norm)} & \textbf{31.4} & \textbf{0.000022} & \textbf{Selected} \\
\bottomrule
\end{tabular}
\end{table*}

The need for extensive L1 tuning (spanning two orders of magnitude) reflects
the non-standard statistical properties of delta activations relative to
full residual stream activations. RMS normalisation was critical: without it,
the MSE term dominates and L1 penalties in the standard range ($10^{-4}$--$10^{-2}$)
are insufficient to enforce sparsity.

\subsection{Effect of Epochs on Reconstruction}

We observe that reconstruction error on held-out data stabilises after
approximately 5--7 epochs for most layer/rank combinations. The remaining
error ($\sim$0.34 at layer 5; $\sim$0.61 at layer 38) appears to be an
irreducible component, potentially reflecting a genuinely distributed aspect
of the adapter's activation contributions that resists sparse decomposition.

\subsection{Subspace Dimension Sensitivity}

Principal angle results are robust to the choice of subspace dimension $k$.
Using $k \in \{64, 128, 256, 512\}$ yields mean principal angles within
$\pm 2°$ of the reported values, confirming that the subspace separation is
not an artifact of the specific $k$ chosen.

\section{Discussion}

\subsection{Convergent Evidence for Geometric Separation}

Three independent analyses consistently indicate that LoRA adapter feature
representations are geometrically separated from base model representations:

\begin{enumerate}[leftmargin=*, topsep=2pt, itemsep=2pt]
    \item \textbf{Cosine similarity}: Mean max similarity $\approx 0.071$
    across all 268M comparisons, barely above random ($\approx 0$).
    \item \textbf{Principal angles}: Mean $\approx 74°$ with 0\% aligned
    directions — rigorous subspace-level evidence.
    \item \textbf{CKA}: Minimum CKA$(h_b, h_\Delta) \approx 0.05$ at layer
    18, the key semantic processing layer.
\end{enumerate}

The convergence of three different measures reduces the risk of any single
metric being misleading. Cosine similarity could in principle reflect SAE
training variance; principal angles address this by operating directly on
the full decoder matrices without SAE decomposition; CKA validates at the
activation level rather than the weight level.

\subsection{The Monitoring Gap}

Our findings imply a concrete safety concern: \textit{interpretability
tools trained on base model activations may be systematically blind to
adapter contributions}. If an organisation deploys a safety-fine-tuned
LoRA model and uses Gemma Scope-based analysis to audit it, the SAE's
reconstruction failure ($\varepsilon > 1.0$) means the tool is not
capturing the adapter's representational contributions. These results suggest that interpretability tools trained solely on base-model activations may incompletely capture adapter-induced representations.

The delta SAE framework we introduce provides a direct solution: training
adapter-specific SAEs enables meaningful feature-level auditing of
fine-tuned models.

\subsection{Rank Effects and Representational Capacity}

Rank affects feature density (monotonically increasing at deep layers)
and CKA distance (higher rank $\to$ lower CKA at semantic layers) but does
not affect the fundamental geometric novelty ($\sim$74° principal angles
across all ranks). This suggests that rank controls \textit{how many}
weakly aligned features the adapter activates, but not the \textit{nature} of those
features — all ranks produce geometrically separated representations.

\subsection{Limitations}

\textbf{Single seed}: All adapters trained with seed 42; seed stability
analysis is deferred to future work.

\textbf{Single dataset}: Results are based on Alpaca instruction tuning;
generalisation to other fine-tuning objectives (e.g., domain adaptation,
RLHF) requires further study.

\textbf{Undertrained adapters}: Despite loss reduction from $\sim$2.0 to
$\sim$1.0, the adapters produce outputs identical to the base model on test
prompts, suggesting the base model's strong priors dominate. This precludes
causal validation via activation steering and is the primary limitation of
the current work.

\textbf{SAE training scale}: Delta SAEs are trained on $\sim$1M token
vectors. Larger-scale training may improve reconstruction quality and alter
the feature vocabulary.

\textbf{No baseline comparison}: We lack a base-base SAE similarity
baseline (two independently trained SAEs on base model activations), which
would allow direct comparison to assess whether 0.071 cosine similarity is
genuinely low or attributable to SAE training variance. This is a priority
for future work.

\section{Related Work}

\textbf{LoRA and PEFT}: \citet{hu2022lora} introduced LoRA; subsequent work
explored variants \citep{dettmers2024qlora, zhao2024lorafa}. \citet{hu2023llm}
studied behavioural effects of fine-tuning, and \citet{biderman2024lora}
studied forgetting dynamics. None have studied fine-tuning effects at the
feature level.

\textbf{Mechanistic interpretability}: SAEs have emerged as a key tool for
decomposing superposed representations \citep{bricken2023monosemanticity,
templeton2024scaling}. \citet{cunningham2023sparse} trained SAEs on GPT-2;
\citet{gemmascope2024} released comprehensive SAE dictionaries for Gemma-2.
All prior work focuses on base models.

\textbf{Fine-tuning and safety}: \citet{qi2023finetuning} and
\citet{yang2023shadow} demonstrated that safety fine-tuning is fragile to
subsequent fine-tuning. \citet{evans2025emergent} showed that narrow
task fine-tuning on insecure code causes broad emergent misalignment.
Our work provides a mechanistic account at the feature level of why
fine-tuning may escape base model safety constraints.

\textbf{Representation similarity}: CKA \citep{kornblith2019similarity} and
SVCCA \citep{raghu2017svcca} have been used to compare model representations
across architectures and training runs; we apply these to compare adapter
and base model representations within a single model.

\section{Conclusion}

We presented a systematic mechanistic interpretability analysis of
LoRA adapter feature geometry using Sparse Autoencoders. Our delta SAE
framework — training SAEs on adapter-induced activation deltas
$\mathbf{h}_{\Delta} = \mathbf{h}_{\text{adapted}} - \mathbf{h}_{\text{base}}$
— provides a mechanistically clean decomposition of adapter contributions.

Three convergent analyses demonstrate that LoRA adapters operate in a
feature subspace geometrically separated from the base model: cosine
similarity near random (0.071), principal angles consistently near 74°,
and CKA$(h_b, h_\Delta) \approx 0.05$--$0.35$ depending on layer.
Adapter-specific SAEs achieve 46--86\% lower reconstruction error than base
model SAEs on held-out data, confirming the geometric separation reflects
genuine learned structure. Feature density scales with rank while geometric
novelty remains rank-invariant.

These findings identify a monitoring gap in LoRA-based safety alignment:
base-model interpretability tools may be systematically blind to
adapter-encoded representations. The delta SAE framework provides a
practical tool for feature-level auditing of fine-tuned models.

Future work will pursue causal validation using strongly-differentiated
models (gemma-2-9b-it), extend to misalignment detection in the style of
\citet{evans2025emergent}, and establish base-base SAE similarity baselines
to rigorously bound the observed geometric separation.

\begin{figure}[!t]
\centering
\includegraphics[width=\textwidth]{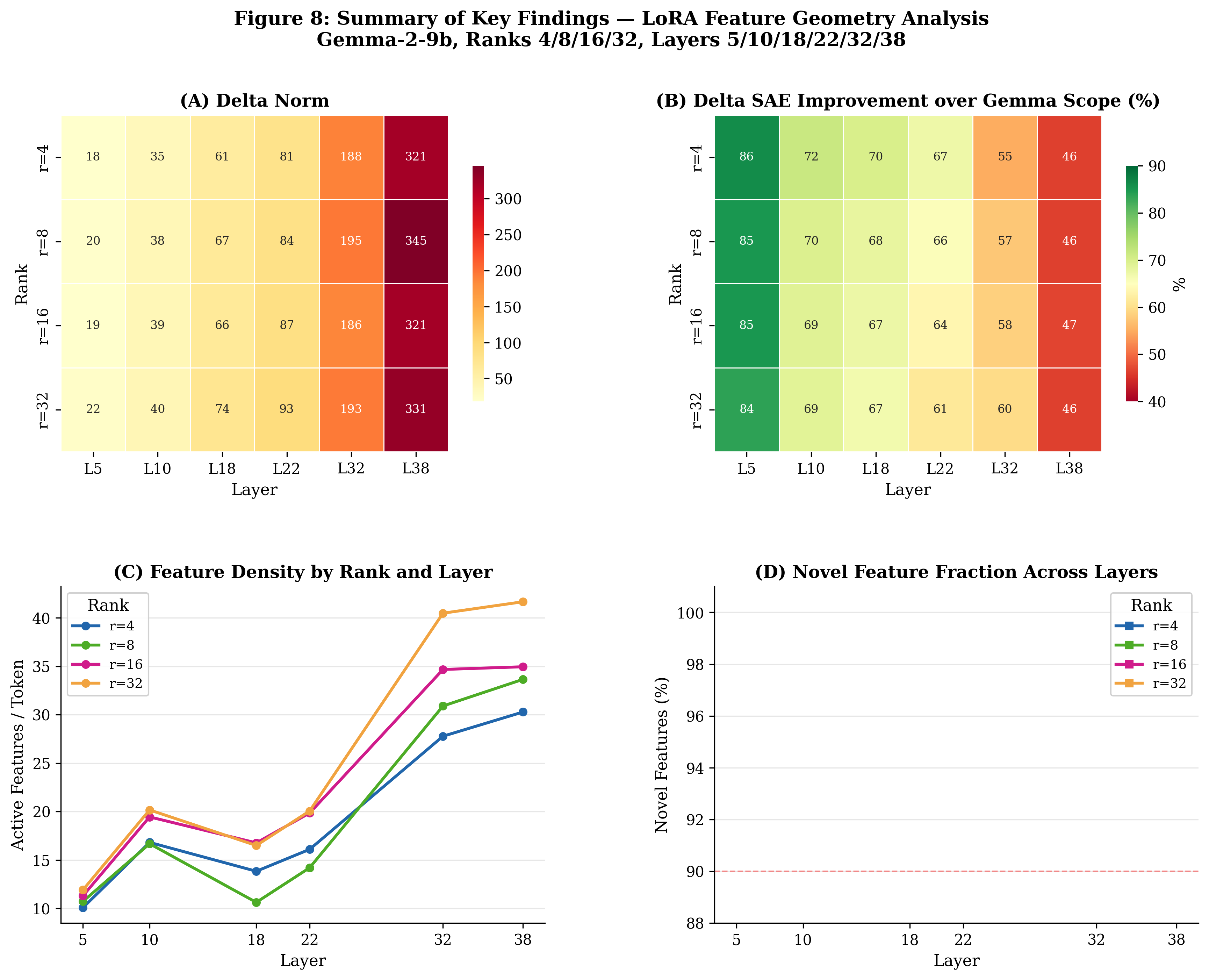}
\caption{Summary of key findings across all ranks and layers.
(A) Delta norm heatmap showing amplification with depth.
(B) Reconstruction improvement of delta SAE over Gemma Scope.
(C) Feature density scaling with rank and depth.
(D) Weakly aligned features fraction — consistently above 93\% across all conditions.}
\label{fig:combined}
\end{figure}

\section*{Acknowledgements}

The author thanks the Gemma Scope team at Google DeepMind for releasing
comprehensive open SAE dictionaries, and the SAELens and TransformerLens
teams for open-source tooling. Experiments were conducted on Apple M4 Studio.
\clearpage

\bibliographystyle{plainnat}

\end{document}